\documentclass[11pt]{article}
\PassOptionsToPackage{hidelinks}{hyperref}

\usepackage[margin=1in]{geometry}
\usepackage[utf8]{inputenc}
\usepackage[T1]{fontenc}
\IfFileExists{lmodern.sty}{\usepackage{lmodern}}{}
\usepackage[expansion=false]{microtype}
\usepackage{amsmath,amssymb}
\usepackage{booktabs}
\usepackage{array}
\usepackage{graphicx}
\usepackage{xcolor}
\usepackage{authblk}
\usepackage{orcidlink}
\usepackage{enumitem}
\usepackage{hyperref}
\usepackage{xurl}
\usepackage[numbers,sort&compress]{natbib}

\newcommand{\pmx}[2]{$#1 \pm #2$}
\newcommand{\brl}{R\$\,}
\newcommand{\releasestatus}{released at \url{https://doi.org/10.5281/zenodo.22948895}}

\title{PixSim: a calibrated open-source simulator of instant-payment fraud, recovery and interdiction under analyst capacity constraints}

\author[1,*]{Bashir Zeimarani\,\orcidlink{0000-0002-7555-8880}}
\author[2]{Alireza Khatib\,\orcidlink{0009-0009-6948-0616}}
\author[2]{Somayeh Mousavinasr\,\orcidlink{0000-0002-4235-6449}}
\author[3]{Carlos Maurício Serodio Figueiredo\,\orcidlink{0000-0002-4484-4411}}
\affil[1]{Instituto CERTI Amazônia, Manaus, Brazil}
\affil[2]{Universidade Federal do Amazonas, Manaus, Brazil}
\affil[3]{Universidade do Estado do Amazonas, Manaus, Brazil}
\affil[*]{Corresponding author: bsz@certi.org.br}

\date{}

\begin{document}
\maketitle

\begin{abstract}
Brazil's Pix settles about 5.9 billion instant, irreversible transfers a month. A fraudulent transfer can be recovered only while the funds remain in a traceable account, and in 2025 the Central Bank's recovery mechanism (MED) returned 9\% of accepted contested value. Interdiction therefore has to happen before settlement, by routing each transaction to pass, human review or block, under a finite analyst team and a regulatory hold window. To our knowledge no public simulator jointly models irreversible settlement, a regulated recovery mechanism, downstream fund dispersal and capacity-constrained review. We present PixSim, an open-source simulator of the Pix rail with these elements, calibrated to Banco Central do Brasil open data: every parameter is sourced, calibrated to one published observable measured under the rules the simulator models, or registered as an assumption. With the model frozen, full-scale runs reproduce the 2025 recovery rate within 0.006 and its decomposition within 0.02; the February--April 2026 window is not reproduced and is reported as a misfit; the layered-tracing regime of May 2026 is a projection. On a benchmark with a payer-side scorer, four reference policies (plus a rank-space variant) and ten scenarios, and within the simulated mule and forwarding model, post-settlement recovery is strongly constrained by dispersal speed; staffing analysts by the arrival profile cuts a fixed rule's alert expiry from 52\% to 2\% at constant hours; halving the team removes the advantage of a fixed threshold-and-block rule over a queue-aware rule, on fraud loss and on loss plus false-block harm (a shift of $+0.106$ of victim value, positive on all twenty paired seeds), while a reversal suggested by an exploratory sweep at two thirds of the team was not confirmed on ten independent seeds; and a synthetic scorer of held-out AUC 0.82 reduces lost value by about a quarter relative to no intervention. Code, calibration data, a parameter register and the summary results are \releasestatus.
\end{abstract}

\section{Introduction}
\label{sec:intro}

Pix, the instant payment rail operated by Banco Central do Brasil (BCB), processed 5.94~billion transactions worth \brl{}2.47~trillion per month in 2025, mean ticket \brl{}414~\cite{bcb_olinda}. Settlement completes in seconds and is final: the paying institution cannot reverse a settled transfer. The recourse is the \emph{Mecanismo Especial de Devolução} (MED): the victim's institution reports the fraud, the receiving institution blocks whatever balance remains, and the blocked amount is returned after analysis~\cite{bcb_res1,bcb_med_guide}. This works only while the money is still there. In 2025, 52.3\% of accepted contested value was not returned for lack of balance and 29.8\% remained after a partial return, 82\% between them, and the MED returned 8.96\%~\cite{bcb_olinda}. A case transcribed from a BCB training deck shows why: five victims, four first-layer mules forwarding two to four minutes after credit, one hub, about twenty third-layer accounts, \brl{}863{,}000 moved through four hops in under five hours~\cite{bcb_med2_deck}. With forwarding in minutes and inter-institution notification taking one to three minutes~\cite{bcb_tempos}, a block that reaches only the first account catches what the ring has not yet moved.

Fraud control on an instant rail therefore has to act before settlement. Each outgoing transaction receives one of three actions: \emph{pass}, which settles it irreversibly; \emph{alert}, which holds it for human review inside a regulatory window of 30~minutes by day and 60 at night~\cite{bcb_tempos}, after which it settles by default; or \emph{block}, which refuses it and harms a legitimate customer when wrong. Analysts are a finite resource whose capacity varies over the day, and a fraud that passes may still be recovered within a random window set by dispersal speed against reporting delay: a continuous-time control problem with an irreversible action, a stochastic recovery clock and a capacity state.

The closest literature, learning to defer under capacity constraints~\cite{alves2024deccaf,alves2025fifar,mozannar2020consistent,reid2024budgeted}, treats capacity as a state but in discrete time, with a binary decision, no irreversible action and no recovery window. Admission control with deadlines~\cite{stidham1985optimal,sennott1999stochastic} and restless bandits~\cite{whittle1988restless} supply the control-theoretic tools and have not been applied to payments. And there is no testbed: PaySim~\cite{lopezrojas2016paysim}, AMLworld~\cite{altman2023amlworld} and CardSim~\cite{cardsim2025} model neither irreversible settlement, nor a regulated recovery window, nor an analyst queue; academic work on Pix fraud is scarce and its most complete treatment, a taxonomy~\cite{unipampa2025taxonomy}, calls for datasets and simulators; and the BCB publishes aggregate MED statistics but no transaction-level data.

This paper provides the testbed. Our contributions are:
\begin{enumerate}[leftmargin=*,itemsep=2pt]
  \item \textbf{PixSim}, an open-source simulator of the Pix rail with irreversible settlement, the MED clock under the three regimes of 2025--2026, the DICT key directory, mule rings reconstructed from a BCB case, and an analyst queue with time-varying staffing and a regulatory hold window (Section~\ref{sec:simulator}).
  \item \textbf{Calibration and regime-structured validation}: every parameter is sourced, calibrated to a single observable measured under the rules the simulator models, or registered as an assumption; four parameters are calibrated and verified on held-out seeds; with the model frozen, full-scale runs reproduce the 2025 recovery rate and its decomposition within stated deviations, and the February--April 2026 window is reported as a misfit (Sections~\ref{sec:calibration} and~\ref{sec:validation}).
  \item \textbf{A benchmark and its findings}: a payer-side scorer, four reference policies and a rank-space variant, ten scenarios, a value waterfall that separates interdiction, hop holds and recovery routes, and paired statistics on shared seeds; within the calibrated model, recovery is strongly constrained by dispersal speed, the relative performance of the tested rules depends on analyst capacity, and fixed rules fail through score saturation below a held-out AUC of about 0.9 (Sections~\ref{sec:benchmark} and~\ref{sec:results}).
\end{enumerate}

\paragraph{Scope.} PixSim is valid for \emph{payer-side interdiction}, decisions taken by the paying institution on outgoing transactions. Receiving-side mule detection is out of scope because simulated receiver accounts are far more separable than real ones; the receiving institution's cautelary blocks are exogenous; and the policy is applied rail-wide, which corresponds to universal adoption. No public figure anchors how much value real payer-side scoring prevents, so absolute loss levels at the reference scorer are optimistic, and every policy is reported as a curve against scorer quality (Section~\ref{sec:limitations}).

\section{Background: Pix, DICT and the MED}
\label{sec:background}

\paragraph{The rail.} Pix, launched in November 2020, is regulated by Resolução BCB~1 and its consolidated Regulamento~\cite{bcb_res1}. A payer initiates a transfer by key, QR code or manual entry; the paying institution (PSP) debits the payer, the receiving PSP credits the payee, and settlement clears through the BCB in seconds. Transactions are classified by the nature of the parties (P2P, P2B, B2P, B2B); in 2025, P2P and P2B were 44\% and 43\% of transactions by count~\cite{bcb_olinda}. Resolução BCB~142/2021 caps transfers from individual accounts at \brl{}1{,}000 between 20:00 and 06:00 unless the customer has raised the limit~\cite{bcb_res142}.

\paragraph{The DICT.} Keys resolve through the \emph{Diretório de Identificadores de Contas Transacionais}. Before a transfer, the paying PSP queries the DICT and receives, with the account identifiers, statistics on the receiving account: type, age, settlements in the last 90~days and 12~months, fraud markers registered by other institutions, and open reports~\cite{bcb_dict_manual}. These are the only receiver-side fields available to the payer at decision time. Since Resolução BCB~506 (30~September 2025~\cite{bcb_res506}) a participant may mark its own client's CPF or CNPJ on a specific transaction involved in fraud (art.~78-HA), and the participant that created the marking or accepted an infraction notification must reject every Pix of that user and account, as payer or receiver, except returns (art.~89, \S2). Resolução BCB~587 (18~September 2026~\cite{bcb_res587}) recast the marking as founded suspicion of fraud, made the marking participant responsible for it, gave the user a contest procedure with a decision within seven days (art.~78-HC), and wrote into the rules the five-year duration the DICT already applied (art.~78-HD). None of this is modelled: the simulator's markers are key-level, never cancelled, and act only through the payer's score.

\paragraph{The MED and its regimes.} A transaction may be contested within 80~calendar days (30 for a return)~\cite{bcb_res1}. The receiving PSP blocks the contested amount up to the available balance, analyses the case, and returns within deadlines that the rules set~\cite{bcb_med_guide,bcb_tempos}. Independently, a receiving PSP may apply a precautionary block (\emph{bloqueio cautelar}, called a \emph{cautelary block} below) for up to 72~hours on a credit it suspects (Regulamento art.~39-B, §4), and later credits to the root account are retained toward the requested value (art.~41-D, in force since 2023). Table~\ref{tab:medregimes} summarises the three regimes the simulator distinguishes. None is a clean operational configuration: in 2025, self-service contestation became mandatory on 1~October and MED~2.0 flows could be adopted from 23~November, and from 30~September Resolução BCB~506~\cite{bcb_res506} required participants to reject every Pix of a user they had marked or reported, so the last quarter mixes configurations (Section~\ref{sec:calibration} checks that this does not move the calibration targets); the enforcement waiver until 10~May~2026 let institutions adopt the new flows at their own pace, so February--April 2026 mixes old and new flows in unknown proportions; and the BCB stated that tracing production started with a minority of cases and ramped up, with the production graph depth unpublished.

\begin{table}[htbp]
\centering\scriptsize
\caption{The MED regimes distinguished by PixSim. Service levels are the regulatory ones; inter-PSP notification latency is one to three minutes in every regime~\cite{bcb_tempos}.}
\label{tab:medregimes}
\setlength{\tabcolsep}{3pt}
\begin{tabular}{>{\raggedright\arraybackslash}p{2.1cm}>{\raggedright\arraybackslash}p{3.7cm}>{\raggedright\arraybackslash}p{4.1cm}>{\raggedright\arraybackslash}p{4.1cm}}
\toprule
& MED~1.0 & MED~2.0 without tracing & MED~2.0 with tracing \\
\midrule
Window & 2025 & 2~Feb--10~May 2026 & from 11~May 2026 \\
Legal basis & Res.\ BCB~1, Regulamento~\cite{bcb_res1} & Res.\ BCB~493/2025~\cite{bcb_res493}; optional from 23~Nov 2025, enforcement waiver to 10~May & Res.\ 493/2025; top-up until case closure from 1~Jul (Res.\ 559/2026~\cite{bcb_res559}) \\
Complaint opening & Institution channel, 1{,}800~s at p95 & Self-service in the paying app, 300~s at p99 & As without tracing \\
Block execution & After analysis, 24~h at p95 & On registration of the infraction notification & As without tracing \\
Accounts blocked & Root only & Root only & Root and the accounts it forwarded to \\
Returns & Regulamento deadlines & 6~h at p99 & 6~h at p99, sequential along the traced path \\
Cautelary blocks & Individual accounts & Extended to legal entities & As without tracing \\
Role in PixSim & Calibrated baseline & Stated misfit, never fitted & Projection at assumed depth \\
\bottomrule
\end{tabular}
\vspace{2pt}
\begin{minipage}{0.97\linewidth}\scriptsize Not modelled in any column: the rejection of all Pix of a marked or reported user by the responsible participant (Res.~506, from 30~September 2025), and the contest procedure of Res.~587 (from 18~September 2026).\end{minipage}
\end{table}

\paragraph{The hold window.} A paying PSP may hold a suspicious outgoing transaction for up to 30~minutes by day and 60 at night before executing or refusing it~\cite{bcb_tempos}. This is the decision deadline for human review in PixSim: an unresolved alert settles by default.

\paragraph{What the BCB publishes.} The Olinda open-data API exposes monthly series on Pix transactions by nature, payer type and initiation form, on DICT key stock, and on MED contestations since January 2022: counts and values contested, accepted and returned, the latter decomposed into full and partial returns, not returned for lack of balance, account closed, and other~\cite{bcb_olinda}. At the time of writing the MED series ends in April 2026, so the tracing regime has no published statistic. Fraud-type shares are not published. Every series and document used is archived with its retrieval date.

\section{Related work}
\label{sec:related}

\paragraph{Learning to defer under capacity.} Learning to defer~\cite{mozannar2020consistent} decides per instance whether to predict or hand off to a human. DeCCaF~\cite{alves2024deccaf} adds per-expert workload constraints, FiFAR and OpenL2D~\cite{alves2025fifar} supply a synthetic fraud setting with simulated experts, budgeted online deferral~\cite{reid2024budgeted} treats the human budget sequentially, and FALCON~\cite{falcon2026} models analyst fatigue as a constrained MDP. Capacity as a state is therefore prior art. All of these are discrete-time, batch or per-round formulations with a binary decision; none has an action that is irreversible and harmful when wrong, a decision deadline, or a recovery window that depends on adversary behaviour. PixSim makes those three features the object of study.

\paragraph{Stochastic control.} The three-action problem with finite review capacity is an admission-control problem with deadlines: alerts are jobs that abandon (settle) if not served within the hold window. Threshold optimality for admission control~\cite{stidham1985optimal,sennott1999stochastic} and index policies for restless bandits~\cite{whittle1988restless} are the natural tools; neither has been applied to fraud review on a payment rail. The benchmark here is built to test policies derived from them; the derivation is a companion paper.

\paragraph{Risk control under drift.} Adaptive conformal inference~\cite{gibbs2021adaptive} and conformal risk control~\cite{angelopoulos2024conformal} bound a monotone loss under exchangeability, and non-exchangeable variants~\cite{farinhas2024nonexchangeable} handle drift by weighting. A control policy breaks exchangeability in a specific way, since a blocked transaction is never observed as fraud or not; PixSim records ground truth for every transaction regardless of action, so it can test guarantees under this censoring.

\paragraph{Simulators and Pix.} PaySim~\cite{lopezrojas2016paysim} and AMLworld~\cite{altman2023amlworld} are agent-based generators of transaction logs with injected fraud or laundering, and CardSim~\cite{cardsim2025} is a card-fraud simulator calibrated to public statistics; none models settlement finality, a regulatory recovery mechanism, a review queue, or a policy interface acting on the flow. On Pix, the most complete academic treatment we found is a taxonomy~\cite{unipampa2025taxonomy} that names the absence of datasets and simulators as the main obstacle; the BCB's MED statistics~\cite{bcb_olinda} and MED~2.0 training material~\cite{bcb_med2_deck} are used here for calibration for the first time. Table~\ref{tab:related} compares the mechanisms implemented by PixSim and its closest alternatives.

\begin{table}[htbp]
\centering\scriptsize
\caption{Mechanisms implemented by PixSim and the closest simulators and human-review frameworks, as described in the cited papers. \checkmark: implemented; --: not implemented. $^{*}$Generated datasets are public; public availability of the generator's source code was not verified.}
\label{tab:related}
\setlength{\tabcolsep}{3pt}
\begin{tabular}{lccccccc}
\toprule
& Public-data & Irreversible & Regulated & Multi-hop & Analyst & Policy acts & Open \\
& calibration & settlement & recovery & dispersal & capacity & on the flow & code \\
\midrule
PaySim~\cite{lopezrojas2016paysim} & -- (private logs) & -- & -- & one hop & -- & -- & \checkmark \\
AMLworld~\cite{altman2023amlworld} & -- & -- & -- & \checkmark & -- & -- & data$^{*}$ \\
CardSim~\cite{cardsim2025} & \checkmark & -- & -- & -- & -- & -- & \checkmark \\
FiFAR / OpenL2D~\cite{alves2025fifar} & -- & -- & -- & -- & \checkmark & -- & \checkmark \\
FALCON~\cite{falcon2026} & -- & -- & -- & -- & \checkmark\ (fatigue) & -- & \checkmark \\
PixSim & \checkmark & \checkmark & \checkmark & \checkmark & \checkmark & \checkmark & \checkmark \\
\bottomrule
\end{tabular}
\end{table}

\section{The simulator}
\label{sec:simulator}

\begin{figure}[htbp]
\centering
\includegraphics[width=\textwidth]{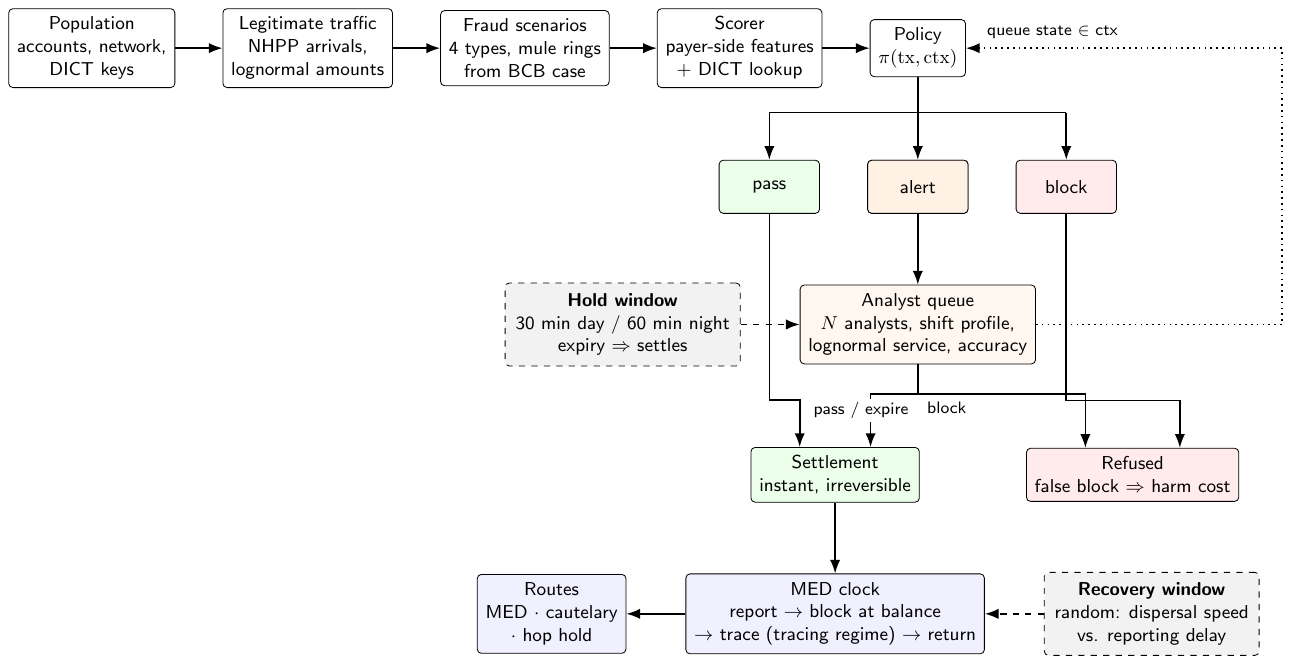}
\caption{The PixSim pipeline. Transactions are scored and routed to pass, alert or block. Alerts wait in a finite analyst queue under the regulatory hold window and settle on expiry. Settled frauds enter the MED clock, whose outcome depends on the random recovery window set by dispersal speed against reporting delay. Queue state is part of the policy's context.}
\label{fig:pipeline}
\end{figure}

Identifiers of the form A1.15 refer to the parameter register described in Section~\ref{sec:calibration}. PixSim is a discrete-event simulator in Python with one module per stage (Figure~\ref{fig:pipeline}). A run builds a population, generates legitimate traffic and fraud over a horizon, passes every transaction through a policy, settles or refuses it, and runs the MED clock of the selected regime on every settled fraud. Every published BCB quantity is recomputed from the simulated log in the BCB's definitions.

\subsection{Population and legitimate traffic}
\label{sec:population}
Accounts are individuals, businesses, mules and exit accounts, each with an opening date, activity rate, device set and region, connected by a preferential-attachment counterparty network with a locality term. A share of legitimate transfers goes to newly created keys, anchored on the monthly DICT key-stock series, and a small share (0.0357, calibrated) is drawn uniformly over all accounts. Transfers arrive as a nonhomogeneous Poisson process whose intensity carries the BCB hourly profile (interpolated from even hours), the weekday shares (Sunday to Saturday 8, 15, 15, 15, 15, 17, 14\%) and a payday spike; account-level independence is a modelling assumption, not an empirical finding. Amounts are lognormal by nature, with $\sigma$ from the published mean and median (P2P 1.81, P2B 1.78, B2P 1.87, B2B 2.60) and $\mu$ solved to the 2025 means. Full scale is 1.08~million accounts and about 35.7~million transactions per 30-day part, about two thirds of them in the 20-day evaluation window. At a 50-thousand-account checkpoint, with nothing tuned, monthly count and value were within 1\% of the BCB series, the mean ticket exact, nature shares within 3 points and weekday shares within 1 point; 36\% of simulated transactions fall under \brl{}20 against 26\% in the BCB's 2020--2022 distribution, a known gap.

\subsection{Fraud scenarios and mule networks}
\label{sec:fraud}
Four scenarios: a social-engineering scam, an account takeover (respecting the night limit), a \emph{Pix errado} refund scam, and QR-code substitution at a merchant; the mix (0.63, 0.21, 0.08, 0.08) has no public source and is registered as an assumption. Victims are sampled with probability proportional to a risk propensity times activity rate. Fraud is drawn at the observed accepted-MED incidence, not at an unknown true prevalence: 0.030\% of transactions are contested and 0.0056\% accepted, with a mean accepted ticket of \brl{}1{,}975~\cite{bcb_olinda}; fraud that is never reported or never reaches the MED is not represented. Rings follow the layered structure of the BCB case, scaled down so that about 45 rings exist at the simulated mule share (A2.10): six first-layer mules, a hub batching in round thousands every 30~minutes (assumed), four third-layer accounts and an aggregator, with cash-outs sent to exit accounts drawn from a separate pool; the first-hop delay has median 3~minutes from the case and is not tuned. Because hub batches mix victims, every fraud transaction carries a table of victim shares (proportional, a convention), so recovery is accounted per victim. Chain depth follows a mixture with mean 2.0 (assumed; Section~\ref{sec:calibration}). A burn rule retires a mule after a MED notification or block with calibrated probability, drawing a replacement from a reserve. Reporting delay is a two-component lognormal mixture (fast share 0.35, medians 20~minutes and 120~hours) anchored on BCB complaint-opening statistics and not tuned; under the 2026 regimes the opening delay follows the new service level (median 63~s). A drift module moves the scenario mix, tickets and dispersal speed over a 90-day horizon.

\subsection{Rail, hold window and MED clock}
\label{sec:rail}
The simulated rail is a single decision maker: one policy is applied to every transaction, mule forwards included, which corresponds to universal adoption of the same rule by all PSPs rather than to one institution acting alone. A policy is any function $\pi(\text{tx},\text{ctx})\in\{\text{pass},\text{alert},\text{block}\}$, where ctx exposes sender history, the receiver's DICT fields, queue state and time. Pass settles instantly and irreversibly. Block never settles and, if the transaction was legitimate, carries a harm of \brl{}50 plus 2\% of the amount (assumed). Alert holds the transaction up to the regulatory window, after which it settles, as the regulation requires. The rail records action, decider (score rule, analyst, expiry) and settlement time for every transaction.

\label{sec:med}
The recovery clock has four components with distinct origins. Two change between regimes, the complaint-opening delay (ii) and block execution (iv); victim discovery (i) and inter-PSP notification latency (iii) are held fixed in every compared configuration. (i) the victim's discovery-and-decision delay, the two-component mixture of Section~\ref{sec:fraud}, anchored on 2025 complaint statistics and held fixed across regimes; (ii) the complaint-opening delay of the paying institution's channel, derived from the regulatory service level of each regime (1{,}800~s at p95 in 2025, 300~s at p99 from 2026) with an assumed lognormal dispersion, so the 2026 median of 63~s is an assumption about the channel and not a claim about victim behaviour; (iii) the inter-PSP notification latency of one to three minutes, sourced and constant; (iv) the receiving PSP's block execution, after an analysis delay under MED~1.0 and immediately on notification from 2026. For every settled fraud the clock of the selected regime runs: the victim reports after (i) and (ii); with calibrated probability the receiving PSP has already applied a cautelary block (an exogenous accounting route, described below); the root account is notified after the sourced latency and blocked up to the requested value, after an analysis delay under MED~1.0 (lognormal, p95 24~h) or immediately under the 2026 regimes; later credits that remain at least 60~minutes (assumed) top up the block; under the tracing regime the accounts the root forwarded to are notified with the same latency and blocked at balance; burns apply; returns run sequentially along the path. A mule forward held or blocked by the policy is not a debit, so its value is still in the root balance when the MED block arrives; that value is reported as its own waterfall row, \emph{held at a hop}, never as victim-transfer interdiction. The row classifies the event history and is not an estimate of incremental recovery, which is read from the paired comparison with pass all on the same world. The rail settles transactions in timestamp order, and MED cases are then processed in report-time order. The cautelary block is an accounting route, not a lock on the rail: it is drawn per victim transfer at case time, the value it covers is returned in full and reserved against the mule's balance so that no later case at the same mule can claim it, and the mule's onward forwards are not re-simulated (A3.4, A3.19). Each unit of victim value is classified by first match in this order: (1) \emph{caught}, if the policy blocks the victim transfer, or an analyst refuses it before the hold expires; (2) otherwise the transfer settles and its onward mule forwards are scored in turn, and a forward held or blocked keeps its value at the mule; (3) \emph{recovered via cautelary block}, if the draw applies; (4) the MED block arrives after (i)--(iv) and captures what is then in the root and, under tracing, the notified accounts, cases at a shared mule being served in report order up to their requested value; (5) what the MED captures because a forward was held is \emph{held at a hop}, what it captures otherwise is \emph{recovered via MED}, and the rest is \emph{lost}. Sender histories and DICT counters are built from settled events strictly before the decision time, with the 12-hour DICT staleness. Deterministic tests check these rules: a forward one second after a cautelary-covered credit neither reduces the cautelary return nor moves the value to the MED route; value reserved by a cautelary block is unavailable to a later case at the same mule; a forward before a credit is charged to the prior balance only; and when two victims share a mule balance the blocks sum to that balance, with requested value equal to recovered plus lost in every case.

\subsection{Analyst queue}
\label{sec:queue}
$N$ analysts, staffed by default in proportion to the arrival profile (flat staffing is a scenario). Service time per case is lognormal with median 6~minutes and $\sigma = 0.6$ (A3.9, assumed; no public source). Review accuracy depends on the fraud type: an analyst refuses a fraudulent alert with probability 0.55 for social engineering, 0.85 for account takeover, 0.50 for \emph{Pix errado} and 0.60 for QR substitution, and wrongly refuses a legitimate alert with probability 0.03 (A3.10, assumed; account takeover carries device signals, scams are authorised by the victim). Discipline is first-come-first-served; an analyst finishes the case in hand when a shift ends, alerts that cross a shift boundary stay in the queue, and alerts unresolved at the horizon are counted as expired if their window has passed and excluded otherwise. Capacity is defined relative to alert volume: the alert threshold is tuned so that the threshold policy's alert stream would occupy five analysts at the mean service time, and the baseline team of six is the size that reviews that stream at 80\% utilisation; scenarios scale the team by a multiplier at the same alert stream, so half capacity is three analysts. Service time and review accuracy are not calibrated, so every capacity result is conditional on them.

\subsection{Engineering and reproducibility}
\label{sec:engineering}
Development runs at 100~thousand accounts with chain-level fraud oversampling; every final table is at full scale and the real fraud rate. A chunked rail processes 30-day segments, verified equivalent to the single-pass rail by test, and a 365-day burn-in before day~0 makes alert volume stationary from the first evaluation day. A 30-day full-scale part with five policies takes about 46~minutes and 17.6~GB at two threads. The tables rest on 155 full-scale parts plus twelve regime validation cells: 135 parts (106 for S1, S2 and S4 to S9, 20 for the capacity sweep S10, 9 for the prevalence sweep S11) ran on one 64-vCPU cloud instance in about 9~hours, and the 20 confirmation parts of Section~\ref{sec:ordering} ran on the development workstation with the same pins; Appendix~\ref{app:runs} lists the configuration, hash, seeds and part count behind every table. The environment is pinned (Python 3.13, numpy 2.5, pandas 3.0, scikit-learn 1.9); a part produced on the instance is field-identical to one produced on the development workstation for the same seed and configuration, apart from timing fields. The package has 78 tests.

\section{Calibration protocol}
\label{sec:calibration}

Four rules. Every parameter is sourced (document and article, or series identifier, recorded beside it), calibrated to exactly one published observable, or registered as an assumption with an identifier in a register released with the code. Every calibration target carries the date window it was measured in and the rules in force during that window, and a parameter may be calibrated only to a target produced under the rules the simulator models for that regime; when a rule changes inside a series, the series is split at the change date and each part is a separate regime. A misfit is first searched for as a missing behaviour and only then as a parameter with an observable of its own; three behaviours entered the model this way (the burn rule, notification latency, legitimate payments to new keys), and no free knob was added. Parameters that share a denominator are fitted jointly.

Calibration runs at development scale (100~thousand accounts, 30 days, fraud oversampled), on six-seed means inside each bisection step; the model is then verified at full scale. Table~\ref{tab:calibrated} lists the calibrated parameters. Three, the uniform receiver share, the cautelary block probability and the active mule share, enter the MED denominator together and are fitted by alternating bisections in two passes; the burn probability is fitted to the reuse ratio. A fifth candidate, the dormant mule activity factor, reached the zero boundary of its search and was fixed there. The 2025 window mixes configurations in its last quarter (Section~\ref{sec:background}), and the BCB's material records a rise in contestations and a fall in the acceptance rate from October. On January--September alone the calibrated shares are within one verification standard error of their full-year values (recovery 0.092 against 0.090, full return 0.058 against 0.057, partial 0.034 against 0.032, cautelary 0.182 against 0.177), so the full year is kept as the calibration window; the mule-reuse anchor is the one quantity that moves (3.32 against 3.19, about four verification standard errors), so on that window alone it would fail the criterion below; it is kept on the full-year value and recorded as a limitation of that anchor. Chain depth is \emph{not} calibrated: the only published statistic that responds to it belongs to the tracing regime, for which no series exists yet, so depth is an assumption (2.0) and the tracing regime a projection. Verification uses six seeds not used in the bisections, and a parameter passes when its target lies within two standard errors of the verification mean (a fixed absolute tolerance would be tighter than the measurement noise at these seed counts). The criterion tests agreement within the chosen model at the seed counts used; it is not a confidence interval for the parameter, and the values are one fitted configuration under the assumed structure, since several combinations of balance, activity, reporting and burning could in principle produce the same aggregates. A local sensitivity of the main findings to the calibrated values is left to the journal version.

\begin{table}[htbp]
\centering
\setlength{\tabcolsep}{3pt}\scriptsize
\caption{Calibrated parameters (register id), the 2025 observable each was fitted to under the MED~1.0 rules, the final bisection bracket, and the six-seed held-out verification (mean, standard error, distance to target in standard errors). Brackets are search resolutions, not confidence intervals. Chain depth is assumed, not calibrated (see text).}
\label{tab:calibrated}
\begin{tabular}{lp{3.0cm}ccccc}
\toprule
Parameter & Target observable (2025) & BCB & Value & Bracket & Verified (SE) & Dist.\ (SE) \\
\midrule
Uniform receiver share (A1.15) & Partial return share & 0.0323 & 0.0357 & [0.0356, 0.0359] & 0.0315 (0.0034) & 0.22 \\
Cautelary block prob.\ (A3.12) & Cautelary route share & 0.1774 & 0.267 & [0.2656, 0.2688] & 0.1736 (0.0060) & 0.63 \\
Active mule share (A3.22) & Full return share & 0.0573 & 0.0830 & [0.0820, 0.0840] & 0.0572 (0.0042) & 0.03 \\
Mule burn probability (A2.21) & Victim transfers per root mule, monthly & 3.19 & 0.168 & [0.164, 0.172] & 3.184 (0.032) & 0.20 \\
Dormant activity factor (A3.17) & (boundary) & -- & 0 & [0, 0.0016] & -- & -- \\
Mean chain depth (A2.12) & none published & -- & 2.0 (assumed) & -- & -- & -- \\
\midrule
Check: MED route recovery & & 0.0896 & & & 0.0887 (0.0056) & 0.16 \\
Check: not returned, all reasons & & 0.613 & & & 0.638 (0.011) & 2.3 \\
\bottomrule
\end{tabular}
\end{table}

Two things in Table~\ref{tab:calibrated} deserve comment. The cautelary probability is fitted to the cautelary route, which the BCB accounts outside the MED rate, so the two routes do not compete for one observable. And the check on unreturned value is against 0.613, not the BCB's no-balance share of 0.523: the BCB decomposes non-returns into no balance (0.523), account closed (0.017) and other (0.073), while the simulator has a single outcome for a case with nothing to block. The model therefore matches the value not returned without representing account closure or the other reasons as distinct causes. Against it the development-scale verification sits 2.3 standard errors high, the one check that fails the two-standard-error criterion, and the full-scale value is within 0.02 (Section~\ref{sec:regime}).

\section{Validation}
\label{sec:validation}

\subsection{Traffic fidelity}
\label{sec:v2}
Validation V2 compares the simulated traffic with 30 published targets, three seeds at full scale. Of the 26 in-sample targets (2025 monthly series, and the 2020--2022 weekday and amount-band profiles), 23 fall within 10\% of the published value; the three outside are amount bands against the 2020--2022 distribution (under \brl{}20: $+37.5\%$; \brl{}100--499.99: $-15.7\%$; \brl{}500 and above: $-27.2\%$), a comparison with an older distribution rather than a measured 2025 discrepancy. Of four out-of-sample targets for January--August 2026, with no parameter changed, the mean ticket ($-4.6\%$) and the individual-payer share ($+1.2\%$) are within 10\%, and monthly count ($-11.7\%$) and value ($-15.8\%$) fall short, as expected of a population calibrated to 2025 volumes. On the fraud side the mean accepted contested ticket is 12.6\% below the BCB's (\brl{}1{,}727 against \brl{}1{,}975; scenario amounts are assumed). Seed noise on traffic targets is below 1\%; traffic does not depend on the MED regime, and the recovery targets are validated in Section~\ref{sec:regime}.

\subsection{The regime test}
\label{sec:regime}

\begin{table}[htbp]
\centering\small
\caption{Three regimes at full scale (1.08~million accounts, every fraud of a 30-day world with its MED case followed to closure, real fraud rate, three seeds, no interdiction, world regenerated per regime; mean~$\pm$~sd), against the BCB statistics of the corresponding window. MED~1.0 is the calibrated regime; MED~2.0 without tracing is a stated misfit, never fitted; MED~2.0 with tracing is a projection at assumed depth 2.0 and MaxHops 2, with no published target. The simulator has one outcome for a case with nothing to block, so its ``not returned'' row is compared with the sum of the BCB's no-balance, account-closed and other categories; the BCB's no-balance share alone is shown beneath it. Columns sum to one.}
\label{tab:regimes}
\setlength{\tabcolsep}{4pt}
\begin{tabular}{lcccccc}
\toprule
& \multicolumn{2}{c}{MED~1.0 (2025)} & \multicolumn{2}{c}{MED~2.0 no tracing (Feb--Apr 2026)} & MED~2.0 tracing \\
Quantity & Simulated & BCB & Simulated & BCB & Simulated \\
\midrule
Recovery rate (MED route) & \pmx{0.096}{0.001} & 0.090 & \pmx{0.114}{0.012} & 0.142 & \pmx{0.237}{0.005} \\
Returned in full           & \pmx{0.060}{0.006} & 0.057 & \pmx{0.067}{0.015} & 0.101 & \pmx{0.097}{0.010} \\
Partial (returned)         & \pmx{0.036}{0.005} & 0.032 & \pmx{0.047}{0.004} & 0.040 & \pmx{0.141}{0.008} \\
Not returned, all reasons  & \pmx{0.595}{0.014} & 0.613 & \pmx{0.584}{0.016} & 0.614 & \pmx{0.407}{0.023} \\
\quad of which no balance  & -- & 0.523 & -- & 0.519 & -- \\
Residual after partial     & \pmx{0.308}{0.015} & 0.298 & \pmx{0.301}{0.008} & 0.256 & \pmx{0.355}{0.018} \\
Cautelary route            & \pmx{0.172}{0.019} & 0.177 & \pmx{0.175}{0.006} & 0.179 & \pmx{0.183}{0.018} \\
Layer block (share)        & 0 & -- & 0 & -- & \pmx{0.095}{0.007} \\
Notified accounts per case & 0 & -- & 0 & -- & \pmx{0.76}{0.03} \\
\bottomrule
\end{tabular}
\end{table}

Table~\ref{tab:regimes} is the central validation. With every parameter frozen at its development-scale value, the full-scale MED~1.0 run gives a recovery rate of 0.096 against the BCB's 0.090: a deviation of $+0.006$, 7\% of the target and six times the three-seed standard deviation, so it is a systematic overshoot rather than seed noise, though it is inside the 0.02 tolerance set for the check and consistent with the development-scale verification ($0.0887 \pm 0.0056$). The tolerance is an application criterion, not a statistical one. Its scale can be read from the value waterfall (Section~\ref{sec:waterfall}) for one specific kind of error, a uniform perturbation $\Delta$ of the MED recovery rate. Such a perturbation acts only on value that reaches the MED net of cautelary returns: about $0.065/0.096 \approx 0.68$ of victim value under pass all and, under an interdicting policy, at most the value neither caught before settlement nor recovered by a cautelary block (0.369 for threshold plus block and 0.409 for capacity-aware in the baseline). At $\Delta = 0.02$ it would therefore move a loss share by at most about 0.014 under pass all and 0.008 under the interdicting policies. This is a scale for the absolute loss levels, not a bound on calibration error in general: a structural error, in mule balances or forwarding behaviour for example, could affect the policies differently, and its effect on the policy comparisons is not bounded here. A perturbation study of those components is left to the journal version (Section~\ref{sec:limitations}). The full and partial return shares and the cautelary route are within seed noise of their targets. On unreturned value, against the sum of the BCB's non-return categories (0.613, Section~\ref{sec:calibration}) the full-scale value of 0.595 is within 0.02, and the residual after partial return within 0.01. Development and full scale agree on every recovery component within 0.03, which is what justifies calibrating at the small scale and verifying at the large one.

Switching on the rules of the February--April 2026 window, with nothing refitted, raises recovery from 0.096 to 0.114, about a third of the observed rise to 0.142, and leaves the full-return share at 0.067 against 0.101 while the cautelary route matches (the worlds are regenerated per regime in Table~\ref{tab:regimes}; on common random numbers, one world per seed shared by the three regimes, the paired effect of the February 2026 rules is $+0.019$ of contested value (per world $+0.011$, $+0.032$, $+0.013$) and that of layered tracing a further $+0.121$ (per world $+0.122$, $+0.108$, $+0.134$), the same sign on all three paired worlds; these are conditional simulation effects under the modelled rules, and the ranges describe variation over generated worlds, not uncertainty about the mechanism's effectiveness in practice). We report this as a misfit and did not fit any parameter to it, for three reasons. First, the enforcement waiver until 10~May means the window mixes MED~1.0 and MED~2.0 flows in unknown proportions, so no single-regime model should match it. Second, every documented rule of the window was tested as a lever, alone and jointly, and none reproduces it: faster victim reporting reaches the recovery rate only at a boundary and doubles the cautelary route (0.38--0.40 against 0.179); immediate blocking raises recovery without raising full returns; partial tracing adoption adds partial, not full, returns. Third, the published series shows that the population of cases changed over the same months, which is a plausible explanation of the residual without being an identified one: between the 2025 monthly mean and the window, accepted value rose 0.9\% while the number of accepted contestations fell 11\% and the number of full returns fell 32\%, the mean accepted ticket rising from \brl{}1{,}975 to \brl{}2{,}243 and the mean full return from \brl{}604 to \brl{}1{,}574. The doubling of the full-return share is therefore fewer, much larger cases returned in full. A simulator with a fixed fraud population cannot produce that, and the aggregates do not say what caused it: self-service contestation, the mix of old and new flows under the waiver, and changes in fraud types or in which cases are accepted are all consistent with them (the BCB's reported figure for February, 13.3\%, agrees with the series~\cite{bcb_lobo2026}). The statistics changed, the frozen model does not reproduce the change, and a composition change is one explanation the series supports.

The tracing regime is a projection: at assumed depth 2.0 and MaxHops 2 it recovers 0.237, with layer blocks contributing 0.095 and 0.76 accounts notified per case. Section~\ref{sec:s8} sweeps both assumptions. Calibration of this regime follows the first published months of the series.

\subsection{Leakage and structure}
\label{sec:leakage}
The scorer sees only what the paying institution sees (V6): payer-side features and DICT lookup fields, with 12-hour staleness. Oversampling preserves ring structure (V11: hub in-degree, tracing density and reuse within 5\% of full scale) but inflates DICT fraud-marker prevalence 50 to 135 times (V13), which is why the scorer is trained at full scale and no development run appears in a table. About 40\% of victim transfers at full scale (57\% under drift) reach a key already carrying a fraud marker (V14); no public figure anchors this.

\section{Benchmark}
\label{sec:benchmark}

\subsection{Scorer}
\label{sec:scorer}
A histogram gradient-boosting classifier (200 iterations, learning rate 0.08, 31 leaves, minimum 50 rows per leaf, L2 1.0, early stopping off) trained on a warm-up period preceding the evaluation window, on payer-side features (amount, hour, channel, device and counterparty novelty, rolling 90-day sender history) and the DICT lookup fields (receiver type, account age, settlement counts over 90~days and 12~months, fraud markers, open reports) with 12-hour staleness. Single-feature AUC on victim transfers is 0.986 for the DICT 12-month settlement counter and above 0.90 for log amount, amount relative to the sender's mean, and counterparty novelty. The DICT counters are retained as a real signal, since exposing them to paying PSPs is why the BCB publishes them; S6 measures the scorer's dependence on them. Over the twenty baseline seeds the scorer trains to AUC $0.990 \pm 0.005$ and scores held-out victim transfers at $0.954 \pm 0.019$.

Early stopping is off. At a positive rate of 0.0056\%, the library's automatic early stopping (a stratified 10\% validation split of about 440 positives, stopping after ten rounds without improvement in log loss) halted training at iteration 10 on six of ten seeds, before the victim features were learned (held-out AUC 0.60--0.78). With a fixed budget the held-out loss plateaus near iteration 130, and 200 is used. Defaults that are safe elsewhere fail silently at this class ratio.

\subsection{Reference policies}
\label{sec:policies}
\emph{Pass all} is the no-intervention baseline. \emph{Budget threshold} alerts above a score cutoff, capped per hour. \emph{Threshold plus block} adds a higher cutoff above which transactions are blocked without review. \emph{Capacity-aware threshold} raises the alert cutoff on the calibrated score as the queue lengthens, with the cutoff capped at 0.999. A rank-space variant of the last, which alerts a shrinking share of each hour's scores and has no cap, is reported in Appendix~\ref{app:rank}: it is more robust to a saturated scorer and worse on average in every other scenario. Every cutoff, budget and queue-response coefficient was set on the warm-up period of development worlds and frozen before the scenarios ran. Any function with the signature of Section~\ref{sec:rail} can be benchmarked.

\subsection{Scenarios}
\label{sec:scenarios}

\begin{table}[htbp]
\centering\small
\caption{Benchmark scenarios. All at full scale (1.08~million accounts, real fraud rate), 30-day parts after a 365-day burn-in of feature history, the first 10 days training the scorer and the remaining 20 forming the evaluation window (drift: 90-day parts, 20-day warm-up, 70-day window); MED~1.0 rules unless stated. Identifiers follow the repository, where S3 is unused.}
\label{tab:scenarios}
\begin{tabular}{lp{0.62\textwidth}c}
\toprule
Scenario & What changes & Seeds \\
\midrule
S1 baseline & Profile staffing, full scorer, team of 6 & 20 \\
S2 drift & Fraud mix, tickets and dispersal speed drift over 90 days & 5 \\
S4 half capacity & Team of 3 & 20 \\
S5 flat staffing & Same analyst hours spread evenly over the day & 5 \\
S6 no DICT & Scorer without the DICT settlement statistics & 10 \\
S7 scorer quality & Full, no DICT, no amount features, logit noise $\sigma \in \{3, 6, 8, 12, 24\}$ & 3 per point \\
S8 tracing projection & MED~2.0 with tracing, MaxHops $\{1,2\}$ $\times$ depth $\{1.5, 2, 3\}$; pass all and threshold plus block & 3 per cell \\
S9 MED~2.0 no tracing & The rules of Feb--Apr 2026 (stated misfit regime) & 10 \\
S10 capacity sweep & Teams of 4 and 5, on the first ten S1 seeds (with S1 and S4: four levels) & 10 each \\
S10c confirmation & Teams of 4 and 5 on ten further seeds, decision rule fixed before the runs (Section~\ref{sec:ordering}) & 10 each \\
S11 prevalence & Accepted-MED incidence at 0.5, 2 and 5 times the 2025 value, team and policy parameters frozen & 3 each \\
\bottomrule
\end{tabular}
\end{table}

Table~\ref{tab:scenarios} lists the scenarios. S1, S2 and S9 vary the environment, S4, S5 and S10 stress capacity, S6 and S7 vary the scorer, S8 sweeps the two assumptions of the tracing regime, and S11 stresses the fraud rate with everything else frozen. S7's noise levels were chosen at development scale so the sweep spans held-out AUC along the noise points from 0.93 to 0.70.

\subsection{Metrics and the value waterfall}
\label{sec:metrics}
Headline metrics are on victim transfers only, so a policy is credited for stopping a victim's payment and not for intercepting a mule's forward. Samples are modest: a full-scale S1 part carries about 1{,}990 fraud events, and the victim value in its evaluation window averages \brl{}3.25~million (\brl{}2.98 to 3.45~million over the twenty parts), an estimated 1{,}900 victim transfers; the standard deviations over seeds reported below reflect this. Frauds are selected by start time in the evaluation window, and their MED cases are followed to closure: a case reported after the part's last day is processed on the world as generated, with the balances left by the last simulated flows and no later top-up credits. About a quarter of window value is reported after the horizon. Splitting the window shows the effect: under pass all the MED route is 0.075 for frauds starting in the first ten days and 0.055 in the last ten, and loss 0.777 against 0.793; under threshold plus block loss is 0.114 against 0.128. The published values average the two halves, so absolute levels carry about one point of horizon effect, shared by every policy on the same world. Every unit of victim value ends in exactly one of five states, summing to one: \emph{caught before settlement}; \emph{held at a hop} (the transfer settled, but a downstream mule forward was held or blocked, leaving the value in a traceable balance for the MED block); \emph{recovered via MED}; \emph{recovered via cautelary block}; \emph{lost}. A test enforces the identity; when a unit of value qualifies for more than one state it is assigned to the earliest event in its history. The MED recovery rate follows the BCB definition: universe, frauds reported inside the contestation window with a settled victim transfer and value left after cautelary returns; numerator, blocked value at root, notified and topped-up accounts capped at the requested value; denominator, requested value. It is therefore a rate on contested value net of cautelary returns, and it is not the waterfall's MED share, which is a fraction of all victim value including what was caught before settlement; the two differ by construction (0.096 against 0.065 under pass all in the baseline) and are never compared. False blocks are recorded with their origin (score rule or analyst; expiry always passes) with count and harm in reais. Alert volume, expiry share, wait and utilisation complete the set, with two operating-point measures: alert precision, the share of alerted transfers that are victim transfers, and victim-value recall at the alert budget, the share of victim value in the window that the policy alerted or blocked before settlement.

Because every policy runs on the same world per seed, comparisons are paired: for each scenario and policy pair we report the mean paired difference in loss share, its standard deviation over seeds and a 95\% percentile bootstrap interval (10{,}000 resamples; the resampling unit is the seed, one paired world), and for the capacity contrast a difference-in-differences on the seeds shared between S1 and S4.

\section{Results}
\label{sec:results}

\subsection{Recovery is constrained by dispersal speed}
\label{sec:dispersal}
Under pass all in the baseline, 0.065 of victim value is recovered through the MED, 0.150 through cautelary blocks, and 0.785 is lost (Section~\ref{sec:waterfall}). This is a share of all victim value, cautelary returns included, and is not directly comparable with the BCB's decomposition of accepted MED value, of which 0.911 was not returned in 2025 (Table~\ref{tab:regimes}). With first-hop forwarding at a median of 3~minutes and notification at 1 to 3~minutes, a root-only block arrives after most of the value has moved. The 2026 rules without tracing raise the MED recovery rate to 0.114 and the tracing projection to 0.237 (Table~\ref{tab:regimes}); on the waterfall's denominator, all victim value, the projection still loses 0.61 to 0.72 under pass all (Section~\ref{sec:s8}). Within the explored assumptions on rings, balances and notification, most of what a paying PSP wants to prevent has to be prevented before settlement.

\subsection{Staffing profile}
\label{sec:staffing}
At fixed analyst hours, moving from flat staffing to staffing proportional to the arrival profile cuts the expiry of threshold plus block's alerts from 0.518 to 0.021 at full scale (Table~\ref{tab:sixconfig}; the development runs that set the capacity definition showed 42\% to 6\%). It requires neither additional analyst hours nor a new scoring model.

\subsection{Policy ordering and the capacity contrast}
\label{sec:ordering}

\begin{table}[htbp]
\centering\footnotesize
\caption{Six-configuration summary on victim transfers, full scale, mean~$\pm$~sd over $n$ seeds. Loss share is the fraction of victim value lost (the caught fraction and the other outcomes are in Table~\ref{tab:waterfall}); expiry is the share of alerts that settled unresolved; loss plus harm adds the false-block harm (\brl{}50 plus 2\% of the amount per refused legitimate transfer) to lost victim value, both in reais, divided by victim value, and is the objective of Section~\ref{sec:conclusion}. Capacity-aware is the probability-space policy; the rank-space variant is in Appendix~\ref{app:rank}.}
\label{tab:sixconfig}
\setlength{\tabcolsep}{3pt}
\begin{tabular}{lcccc}
\toprule
& Pass all & Budget thr. & Thr.\ + block & Capacity-aware \\
\midrule
\multicolumn{5}{l}{\emph{Loss share}} \\
S1 baseline ($n=20$)      & \pmx{0.785}{0.013} & \pmx{0.408}{0.017} & $\mathbf{0.121 \pm 0.025}$ & \pmx{0.221}{0.014} \\
S2 drift ($n=5$)          & \pmx{0.730}{0.009} & \pmx{0.378}{0.006} & $\mathbf{0.110 \pm 0.031}$ & \pmx{0.195}{0.007} \\
S4 half capacity ($n=20$) & \pmx{0.785}{0.013} & \pmx{0.527}{0.015} & \pmx{0.320}{0.055} & $\mathbf{0.314 \pm 0.017}$ \\
S5 flat staffing ($n=5$)  & \pmx{0.783}{0.009} & \pmx{0.434}{0.015} & \pmx{0.271}{0.018} & $\mathbf{0.267 \pm 0.027}$ \\
S6 no DICT ($n=10$)       & \pmx{0.783}{0.010} & \pmx{0.479}{0.020} & $\mathbf{0.149 \pm 0.021}$ & \pmx{0.315}{0.021} \\
S9 MED~2.0 no tracing ($n=10$) & \pmx{0.777}{0.020} & \pmx{0.398}{0.019} & $\mathbf{0.113 \pm 0.020}$ & \pmx{0.204}{0.013} \\
\midrule
\multicolumn{5}{l}{\emph{Alert expiry share}} \\
S1 baseline      & -- & \pmx{0.022}{0.006} & \pmx{0.021}{0.007} & \pmx{0.008}{0.003} \\
S2 drift         & -- & \pmx{0.032}{0.010} & \pmx{0.048}{0.007} & \pmx{0.012}{0.003} \\
S4 half capacity & -- & \pmx{0.340}{0.020} & \pmx{0.772}{0.008} & \pmx{0.116}{0.014} \\
S5 flat staffing & -- & \pmx{0.023}{0.003} & \pmx{0.518}{0.021} & \pmx{0.031}{0.006} \\
S6 no DICT       & -- & \pmx{0.022}{0.005} & \pmx{0.045}{0.018} & \pmx{0.008}{0.003} \\
S9 MED~2.0 no tracing & -- & \pmx{0.021}{0.006} & \pmx{0.027}{0.012} & \pmx{0.008}{0.003} \\
\midrule
\multicolumn{5}{l}{\emph{Loss plus harm (share of victim value)}} \\
S1 baseline      & \pmx{0.785}{0.013} & \pmx{0.413}{0.017} & \pmx{0.131}{0.025} & \pmx{0.228}{0.014} \\
S2 drift         & \pmx{0.730}{0.009} & \pmx{0.383}{0.006} & \pmx{0.120}{0.031} & \pmx{0.203}{0.007} \\
S4 half capacity & \pmx{0.785}{0.013} & \pmx{0.531}{0.015} & \pmx{0.323}{0.055} & \pmx{0.317}{0.017} \\
S5 flat staffing & \pmx{0.783}{0.009} & \pmx{0.440}{0.015} & \pmx{0.276}{0.018} & \pmx{0.272}{0.027} \\
S6 no DICT       & \pmx{0.783}{0.010} & \pmx{0.486}{0.020} & \pmx{0.160}{0.021} & \pmx{0.324}{0.021} \\
S9 MED~2.0 no tracing & \pmx{0.777}{0.020} & \pmx{0.404}{0.019} & \pmx{0.123}{0.020} & \pmx{0.212}{0.013} \\
\bottomrule
\end{tabular}
\end{table}

\begin{table}[htbp]
\centering\small
\caption{Paired loss-share differences between threshold plus block and the capacity-aware policy (negative favours the fixed rule), with 95\% percentile bootstrap intervals (10{,}000 resamples of paired seeds) for $n \ge 10$; $^{\dagger}$for the five-seed scenarios, the range of the five per-seed differences (same sign on all five seeds in S2, signs differ in S5). The last row is the difference-in-differences of S4 against S1 on their twenty shared seeds.}
\label{tab:paired}
\begin{tabular}{lccc}
\toprule
Scenario & $n$ & Paired difference & 95\% interval \\
\midrule
S1 baseline & 20 & $-0.100$ & $[-0.110, -0.091]$ \\
S2 drift & 5 & $-0.086$ & $-0.136$ to $-0.047^{\dagger}$ \\
S6 no DICT & 10 & $-0.166$ & $[-0.184, -0.149]$ \\
S9 MED~2.0 no tracing & 10 & $-0.091$ & $[-0.102, -0.080]$ \\
S4 half capacity & 20 & $+0.006$ & $[-0.019, +0.030]$ \\
S5 flat staffing & 5 & $+0.004$ & $-0.048$ to $+0.021^{\dagger}$ \\
\midrule
S4 minus S1, shared seeds & 20 & $+0.106$ & $[+0.087, +0.124]$ \\
\bottomrule
\end{tabular}
\end{table}

In the baseline (Table~\ref{tab:sixconfig}) the ordering by loss is threshold plus block (0.121), capacity-aware (0.221), budget threshold (0.408), pass all (0.785), preserved under drift, without DICT statistics and under the 2026 rules: with analysts plentiful, the block rule wins because it removes the highest-scoring transactions without spending review time on them. The fixed rule's advantage over the queue-aware rule is $0.100$ of victim value in the baseline, with a paired interval of $[0.091, 0.110]$ (Table~\ref{tab:paired}). On loss plus harm the ordering is the same in every scenario of Table~\ref{tab:sixconfig}: harm is 1 to 8\% of lost value at the reference scorer, and the ordering is unchanged for harm weights from half to ten times the assumed \brl{}50 plus 2\%, so the cost model separates policies only when the scorer is weak (Section~\ref{sec:s7}).

That advantage depends on capacity. At half capacity (S4, twenty seeds) the paired difference is $+0.006$ with an interval spanning zero, and under flat staffing (S5) $+0.004$ with signs differing across the five seeds: the two rules are indistinguishable. The direct test of the capacity effect is the difference-in-differences on the twenty seeds shared between S1 and S4: halving the team shifts the relative performance by $+0.106$ $[0.087, 0.124]$, positive on every seed (range $+0.011$ to $+0.196$); on loss plus harm the same contrast is $+0.103$ $[0.085, 0.122]$. The expiry rows give the mechanism. A fixed threshold ignores the queue, so when the team shrinks or is mismatched to arrivals, 77\% (S4) and 52\% (S5) of its alerts expire unreviewed, while the capacity-aware policy holds expiry at 12\% and 3\% by alerting less; its catch before settlement falls from 0.516 to 0.402 where the fixed rule's falls from 0.565 to 0.244.

The capacity sweep (S10, Table~\ref{tab:capsweep}) was added after the six-versus-three contrast, to locate the change, and is exploratory: it samples four team sizes on the first ten S1 seeds and its intervals are pointwise, with no correction for examining several levels. The paired difference is $-0.089$ with six analysts, $-0.044$ with five, $+0.046$ $[+0.031, +0.062]$ with four and $+0.021$ $[-0.000, +0.043]$ with three, and loss plus harm gives the same signs at every size ($-0.087$, $-0.041$, $+0.046$ $[+0.031, +0.062]$, $+0.021$). Because the four-analyst value suggested a reversal, it was tested on ten further seeds (20260920--20260929) with a decision rule fixed before the runs: the reversal holds if the 95\% interval of the loss-share difference on these seeds lies above zero. It does not: the difference is $+0.017$ $[-0.016, +0.051]$ (loss plus harm $+0.017$ $[-0.017, +0.051]$), so the reversal is not confirmed. Pooled with the original ten seeds, which generated the hypothesis, the estimate is $+0.031$ $[+0.011, +0.050]$; it is reported for completeness and not as confirmation. At five analysts the new seeds agree with the sweep ($-0.066$ $[-0.086, -0.047]$, descriptive). The confirmation seeds had not been used with teams of four or five, but they are the S1 and S4 seeds 11 to 20, whose six- and three-analyst results were known when the four-analyst level was chosen; the seed list and decision rule were recorded in a snapshot before the first part started, and the analysis script's hash the following day before any edit. What the data support is therefore that reduced capacity erodes and then removes the fixed rule's advantage over the queue-aware rule, with the two statistically indistinguishable between three and four analysts, while the precise crossing remains uncertain; no optimal team size is identified, and the twenty-seed six-versus-three contrast addresses a different comparison. These are frozen reference implementations, not optimised representatives of their families: the data support that the relative performance of the tested rules depends on capacity, and the optimal cutoff as a function of queue state is the companion paper's question.

\begin{table}[htbp]
\centering\small
\caption{Capacity sweep (S10 with S1 and S4) on the ten shared seeds 20260910--20260919, exploratory, with the four- and five-analyst results on ten independent confirmation seeds: loss share and loss plus harm by policy and team size, mean~$\pm$~sd, and the paired difference threshold plus block minus capacity-aware with its pointwise 95\% percentile bootstrap interval (10{,}000 resamples of paired seeds). Pass all loses $0.783 \pm 0.010$ at every level, with no harm.}
\label{tab:capsweep}
\setlength{\tabcolsep}{3pt}\footnotesize
\begin{tabular}{lccccc}
\toprule
Team & Budget & Thr.+block & Cap.-aware & Paired diff.\ (95\% CI) & Expiry (T+B / CA) \\
\midrule
\multicolumn{6}{l}{\emph{Loss share}} \\
6 (S1, seeds 1--10) & \pmx{0.406}{0.018} & \pmx{0.131}{0.016} & \pmx{0.220}{0.012} & $-0.089$ $[-0.095, -0.083]$ & 0.020 / 0.008 \\
5      & \pmx{0.420}{0.023} & \pmx{0.181}{0.027} & \pmx{0.224}{0.018} & $-0.044$ $[-0.056, -0.029]$ & 0.190 / 0.014 \\
4      & \pmx{0.440}{0.016} & \pmx{0.290}{0.035} & \pmx{0.244}{0.018} & $+0.046$ $[+0.031, +0.062]$ & 0.587 / 0.031 \\
3 (S4, seeds 1--10) & \pmx{0.529}{0.017} & \pmx{0.341}{0.034} & \pmx{0.319}{0.021} & $+0.021$ $[-0.000, +0.043]$ & 0.772 / 0.117 \\
\midrule
\multicolumn{6}{l}{\emph{Loss plus harm (share of victim value)}} \\
6 (S1, seeds 1--10) & \pmx{0.412}{0.018} & \pmx{0.141}{0.016} & \pmx{0.228}{0.013} & $-0.087$ $[-0.093, -0.080]$ & \\
5      & \pmx{0.426}{0.023} & \pmx{0.190}{0.026} & \pmx{0.231}{0.019} & $-0.041$ $[-0.052, -0.027]$ & \\
4      & \pmx{0.446}{0.016} & \pmx{0.295}{0.036} & \pmx{0.249}{0.018} & $+0.046$ $[+0.031, +0.062]$ & \\
3 (S4, seeds 1--10) & \pmx{0.533}{0.017} & \pmx{0.344}{0.034} & \pmx{0.322}{0.020} & $+0.021$ $[-0.000, +0.044]$ & \\
\midrule
\multicolumn{6}{l}{\emph{Confirmation seeds 20260920--20260929, paired difference}} \\
4, loss share & & & & $+0.017$ $[-0.016, +0.051]$ & \\
4, loss plus harm & & & & $+0.017$ $[-0.017, +0.051]$ & \\
5, loss share & & & & $-0.066$ $[-0.086, -0.047]$ & \\
5, loss plus harm & & & & $-0.064$ $[-0.083, -0.045]$ & \\
\bottomrule
\end{tabular}
\end{table}

\subsection{Where the value goes}
\label{sec:waterfall}

\begin{table}[htbp]
\centering\footnotesize
\caption{Value waterfall on victim transfers, full scale, mean~$\pm$~sd over $n$ seeds; each row sums to one; S2 and S9, whose loss shares are in Table~\ref{tab:sixconfig}, are in the repository. Columns: caught before settlement; held at a hop (the transfer settled but the policy held or blocked its onward mule forward, leaving the value for the MED block); recovered via MED; recovered via cautelary block; lost.}
\label{tab:waterfall}
\setlength{\tabcolsep}{3pt}
\begin{tabular}{llccccc}
\toprule
Scenario & Policy & Caught & Hop held & MED & Cautelary & Lost \\
\midrule
S1 ($n=20$) & pass all          & 0 & 0 & \pmx{0.065}{0.010} & \pmx{0.150}{0.009} & \pmx{0.785}{0.013} \\
            & budget threshold  & \pmx{0.324}{0.019} & \pmx{0.128}{0.013} & \pmx{0.038}{0.005} & \pmx{0.103}{0.011} & \pmx{0.408}{0.017} \\
            & threshold + block & \pmx{0.565}{0.045} & \pmx{0.230}{0.030} & \pmx{0.019}{0.005} & \pmx{0.066}{0.013} & \pmx{0.121}{0.025} \\
            & capacity-aware    & \pmx{0.516}{0.030} & \pmx{0.163}{0.018} & \pmx{0.025}{0.005} & \pmx{0.075}{0.009} & \pmx{0.221}{0.014} \\
\midrule
S4 ($n=20$) & pass all          & 0 & 0 & \pmx{0.065}{0.010} & \pmx{0.150}{0.009} & \pmx{0.785}{0.013} \\
            & budget threshold  & \pmx{0.222}{0.016} & \pmx{0.084}{0.010} & \pmx{0.054}{0.008} & \pmx{0.114}{0.013} & \pmx{0.527}{0.015} \\
            & threshold + block & \pmx{0.244}{0.066} & \pmx{0.274}{0.061} & \pmx{0.050}{0.009} & \pmx{0.113}{0.019} & \pmx{0.320}{0.055} \\
            & capacity-aware    & \pmx{0.402}{0.029} & \pmx{0.161}{0.025} & \pmx{0.032}{0.006} & \pmx{0.091}{0.014} & \pmx{0.314}{0.017} \\
\midrule
S5 ($n=5$)  & pass all          & 0 & 0 & \pmx{0.065}{0.010} & \pmx{0.153}{0.007} & \pmx{0.783}{0.009} \\
            & budget threshold  & \pmx{0.312}{0.023} & \pmx{0.120}{0.007} & \pmx{0.040}{0.009} & \pmx{0.093}{0.009} & \pmx{0.434}{0.015} \\
            & threshold + block & \pmx{0.356}{0.064} & \pmx{0.241}{0.031} & \pmx{0.038}{0.010} & \pmx{0.094}{0.019} & \pmx{0.271}{0.018} \\
            & capacity-aware    & \pmx{0.447}{0.042} & \pmx{0.182}{0.020} & \pmx{0.030}{0.009} & \pmx{0.076}{0.011} & \pmx{0.267}{0.027} \\
\midrule
S6 ($n=10$) & pass all          & 0 & 0 & \pmx{0.065}{0.010} & \pmx{0.152}{0.007} & \pmx{0.783}{0.010} \\
            & budget threshold  & \pmx{0.184}{0.017} & \pmx{0.165}{0.017} & \pmx{0.045}{0.006} & \pmx{0.127}{0.011} & \pmx{0.479}{0.020} \\
            & threshold + block & \pmx{0.297}{0.023} & \pmx{0.417}{0.019} & \pmx{0.031}{0.007} & \pmx{0.106}{0.009} & \pmx{0.149}{0.021} \\
            & capacity-aware    & \pmx{0.256}{0.021} & \pmx{0.288}{0.017} & \pmx{0.035}{0.006} & \pmx{0.106}{0.011} & \pmx{0.315}{0.021} \\
\bottomrule
\end{tabular}
\end{table}

Genuine post-settlement recovery falls under interdiction (Table~\ref{tab:waterfall}): from 0.065 under pass all to 0.019 under threshold plus block in the baseline, because the frauds the policy stops are those the MED would otherwise partly recover. What rises is \emph{held at a hop}: 0.230 under threshold plus block, 0.163 under capacity-aware, and 0.417 without DICT statistics (S6). In S6 the scorer catches 0.297 before settlement against 0.565 in S1, yet loss rises only from 0.121 to 0.149 because hop holds rise from 0.230 to 0.417. Hop holds exist because the policy acts on every institution's mule forwards and because simulated forwards may be easier to score than real ones (Section~\ref{sec:limitations}); they are a property of universal adoption and of the simulated forwarding structure, not an estimate for one institution, and claims about payer-side scoring alone should be read off the \emph{caught} column. That column also refines the capacity result: at half capacity the fixed rule catches 0.244 at the transfer and 0.274 at a hop, the capacity-aware rule 0.402 and 0.161 (under flat staffing 0.356/0.241 against 0.447/0.182), so the policy that respects the queue does its work earlier in the chain.

\subsection{Scorer quality}
\label{sec:s7}

\begin{table}[htbp]
\centering\footnotesize
\caption{Scenario S7, scorer-quality sweep, full scale, three seeds per point, ordered by held-out AUC on victim transfers in the evaluation window (mean~$\pm$~sd). The first block gives the operating point that the policies actually act on: precision among alerted victim transfers and the share of victim value alerted or blocked before settlement, under threshold plus block (Section~\ref{sec:metrics}). Pass all loses $0.782 \pm 0.006$ at every point, with no harm. Loss is mean~$\pm$~sd; the loss-plus-harm, expiry and false-block blocks are means. Loss plus harm is the objective of Section~\ref{sec:conclusion}; it exceeds one where false-block harm exceeds the victim value at stake. Both capacity-aware definitions are shown because the sweep is where they differ.}
\label{tab:s7}
\setlength{\tabcolsep}{2.5pt}
\begin{tabular}{lccccc}
\toprule
Point & AUC$_\text{held-out}$ & Budget & Thr.+block & Cap.\ (prob.) & Cap.\ (rank) \\
\midrule
\multicolumn{6}{l}{\emph{Operating point of threshold plus block: alert precision on victim transfers / victim-value recall at the alert budget}} \\
no amount features & \pmx{0.994}{0.003} & \multicolumn{4}{l}{\pmx{0.058}{0.008} / \pmx{0.66}{0.04}} \\
no DICT statistics & \pmx{0.992}{0.001} & \multicolumn{4}{l}{\pmx{0.028}{0.001} / \pmx{0.43}{0.02}} \\
full scorer        & \pmx{0.941}{0.020} & \multicolumn{4}{l}{\pmx{0.063}{0.005} / \pmx{0.88}{0.06}} \\
noise $\sigma=3$   & \pmx{0.934}{0.023} & \multicolumn{4}{l}{\pmx{0.050}{0.003} / \pmx{0.69}{0.05}} \\
noise $\sigma=6$   & \pmx{0.902}{0.027} & \multicolumn{4}{l}{\pmx{0.025}{0.002} / \pmx{0.41}{0.10}} \\
noise $\sigma=8$   & \pmx{0.874}{0.030} & \multicolumn{4}{l}{\pmx{0.016}{0.004} / \pmx{0.30}{0.10}} \\
noise $\sigma=12$  & \pmx{0.818}{0.033} & \multicolumn{4}{l}{\pmx{0.008}{0.003} / \pmx{0.13}{0.05}} \\
noise $\sigma=24$  & \pmx{0.698}{0.028} & \multicolumn{4}{l}{$0.000$ / \pmx{0.12}{0.03} (both cutoffs at the cap: every transaction at the cap is blocked, none alerted)} \\
\midrule
\multicolumn{6}{l}{\emph{Loss share}} \\
no amount features & \pmx{0.994}{0.003} & \pmx{0.445}{0.008} & \pmx{0.149}{0.016} & \pmx{0.271}{0.006} & \pmx{0.277}{0.029} \\
no DICT statistics & \pmx{0.992}{0.001} & \pmx{0.474}{0.027} & \pmx{0.141}{0.011} & \pmx{0.312}{0.018} & \pmx{0.315}{0.006} \\
full scorer        & \pmx{0.941}{0.020} & \pmx{0.404}{0.018} & \pmx{0.122}{0.012} & \pmx{0.212}{0.010} & \pmx{0.223}{0.011} \\
noise $\sigma=3$   & \pmx{0.934}{0.023} & \pmx{0.434}{0.011} & \pmx{0.145}{0.024} & \pmx{0.260}{0.013} & \pmx{0.284}{0.024} \\
noise $\sigma=6$   & \pmx{0.902}{0.027} & \pmx{0.487}{0.016} & \pmx{0.198}{0.030} & \pmx{0.332}{0.018} & \pmx{0.340}{0.033} \\
noise $\sigma=8$   & \pmx{0.874}{0.030} & \pmx{0.546}{0.003} & \pmx{0.312}{0.027} & \pmx{0.746}{0.012} & \pmx{0.407}{0.018} \\
noise $\sigma=12$  & \pmx{0.818}{0.033} & \pmx{0.675}{0.016} & \pmx{0.574}{0.004} & \pmx{0.759}{0.005} & \pmx{0.615}{0.007} \\
noise $\sigma=24$  & \pmx{0.698}{0.028} & \pmx{0.692}{0.014} & \pmx{0.549}{0.010} & \pmx{0.742}{0.017} & \pmx{0.760}{0.016} \\
\midrule
\multicolumn{6}{l}{\emph{Loss plus harm (share of victim value)}} \\
no amount features & & 0.449 & 0.156 & 0.276 & 0.282 \\
no DICT statistics & & 0.480 & 0.152 & 0.321 & 0.324 \\
full scorer        & & 0.410 & 0.132 & 0.219 & 0.231 \\
noise $\sigma=3$   & & 0.439 & 0.155 & 0.267 & 0.292 \\
noise $\sigma=6$   & & 0.493 & 0.207 & 0.339 & 0.347 \\
noise $\sigma=8$   & & 0.552 & 0.326 & 0.746 & 0.415 \\
noise $\sigma=12$  & & 0.682 & 0.607 & 0.759 & 0.624 \\
noise $\sigma=24$  & & 0.845 & 7.81 & 0.747 & 0.819 \\
\midrule
\multicolumn{6}{l}{\emph{Alert expiry share}} \\
no amount features & & 0.020 & 0.033 & 0.010 & 0.055 \\
no DICT statistics & & 0.020 & 0.044 & 0.007 & 0.047 \\
full scorer        & & 0.020 & 0.019 & 0.010 & 0.039 \\
noise $\sigma=3$   & & 0.022 & 0.020 & 0.018 & 0.028 \\
noise $\sigma=6$   & & 0.006 & 0.013 & 0.012 & 0.016 \\
noise $\sigma=8$   & & 0.009 & 0.011 & 0.992 & 0.010 \\
noise $\sigma=12$  & & 0.008 & 0.007 & 0.999 & 0.004 \\
noise $\sigma=24$  & & 0.002 & 0.000 & 0.998 & 0.874 \\
\midrule
\multicolumn{6}{l}{\emph{False blocks (count)}} \\
no amount features & & 227 & 386 & 306 & 343 \\
no DICT statistics & & 244 & 418 & 338 & 372 \\
full scorer        & & 238 & 407 & 296 & 339 \\
noise $\sigma=3$   & & 264 & 446 & 351 & 372 \\
noise $\sigma=6$   & & 316 & 489 & 415 & 419 \\
noise $\sigma=8$   & & 338 & 796 & 29 & 458 \\
noise $\sigma=12$  & & 385 & 1{,}954 & 18 & 521 \\
noise $\sigma=24$  & & 8{,}894 & 422{,}800 & 321 & 3{,}451 \\
\bottomrule
\end{tabular}
\end{table}

Table~\ref{tab:s7} sweeps eight scorers along held-out victim-transfer AUC in the evaluation window (the full-scorer point uses three of the S1 seeds; with three seeds per point, small differences between adjacent points are not resolved, and paired differences are summarised by their sign across the three seeds rather than by an interval). Threshold plus block has a lower loss than the capacity-aware policy at every point, on all three seeds, and the lowest loss of all rules, budget threshold degrades least (+0.29 at worst), and the capacity-aware policy sits between them down to AUC 0.90. Below about 0.9 each fixed rule fails in its own way, and each failure is a saturated score. The probability-space capacity-aware policy collapses from $\sigma=8$, where its threshold reaches the 0.999 cap and every score above it is alerted regardless of the queue (expiry 0.99, and at most 0.005 of victim value caught before settlement); the rank-space variant holds to $\sigma=12$ and collapses at $\sigma=24$ (expiry 0.87); threshold plus block fails through false blocks, since the block quantile among saturated scores is 1.0 and every score at the ceiling is refused: 1{,}954 false blocks at $\sigma=12$, and 423~thousand, about 24~million reais of harm, at $\sigma=24$. The loss column hides this and the loss-plus-harm rows show it: at $\sigma=24$ threshold plus block's total is 7.8 times the victim value at stake and budget threshold's 0.845 exceeds pass all's 0.782, so the best rule at the weakest point is no intervention; at $\sigma=12$ threshold plus block has the lowest total at the assumed harm weight (0.607 against 0.624 for the rank-space rule and 0.682 for budget threshold), a margin the three-seed means cannot confirm at twice the weight (0.640 against 0.633). For the evaluated synthetic scorer and threshold plus block, the point at held-out AUC 0.82 leaves 0.574 lost against 0.782, a 27\% reduction relative to no intervention (22\% on loss plus harm); this does not imply that another scorer of the same AUC, real or synthetic, would do as well. No public figure places real payer-side scorers on this axis, so absolute levels should be read over this range rather than at the baseline's 0.121.

\emph{The axis is meaningful along the noise points, not across feature sets.} Removing the DICT statistics or the amount features raises held-out victim-transfer AUC to 0.99 while worsening every policy's loss. The features do not act on other rows (on non-victim rows the three feature sets are within 0.01 of each other in AUC); what they change is the top tail of the score, where the policies act, while a global AUC over millions of legitimate rows at a positive rate of 0.0056\% is nearly saturated. Under threshold plus block the full scorer has alert precision on victims of 0.063 against 0.028 without DICT statistics, and victim-value recall at the alert budget of 0.88 against 0.43 (first block of Table~\ref{tab:s7}); along the noise points both fall monotonically with AUC, from 0.88 recall to 0.13 at $\sigma=12$, which is the quantity the loss column follows. Feature-set points are therefore compared on precision, recall at budget and loss, and held-out AUC is used only along the noise points, where the scorer's inputs do not change.

\subsection{The tracing projection}
\label{sec:s8}
Scenario S8 sweeps the two assumptions of the tracing regime with no interdiction and with threshold plus block. Under pass all, lost value falls from 0.717 (MaxHops 1, depth 1.5) to 0.606 (MaxHops 2, depth 3.0) and the MED route's share rises from 0.139 to 0.248: deeper chains leave more value in traceable intermediate accounts, and a second notified layer reaches it. Under threshold plus block, loss is 0.083 to 0.095 across the grid: interdiction before settlement dominates and the tracing assumptions barely matter to it. The regime's calibration waits on the first published months of the series; until then S8 gives the sensitivity range of what tracing adds under the two assumptions, not an empirical or analytical bound. The tracing cells also carry a small overstatement: because the cautelary route reserves only the root's balance and forwards are not re-simulated, a layer block can capture value that a cautelary return, usually the same case's partial one, already counts as returned. Measured by an instrumented run of the same clock at a quarter of full scale (one seed), the double-counted value is 0.008 to 0.071 of layer recoveries across the S8 grid (0.003 to 0.019 of MED-route recoveries, at most 0.005 of contested value), rising with depth; the regimes without tracing block at the root only and are unaffected.

\subsection{Fraud prevalence}
\label{sec:s11}
The accepted-MED incidence is a lower bound on true prevalence. S11 draws fraud at 0.5, 2 and 5 times that rate with the team, the scorer's training protocol and every policy parameter frozen (three seeds per level). The ordering of every policy pair is the same at 0.5, 1 and 2 times: threshold plus block loses 0.053, 0.122 and 0.123 of victim value, the capacity-aware policy 0.193, 0.212 and 0.248, and the scorer trains no worse (held-out victim AUC 0.991 at five times, on 22{,}000 positives). At half the rate threshold plus block's loss halves with expiry unchanged, so the gain is not a queue effect: with the same alert budgets and half the fraud, the budget covers more of the fraud value at stake, and the waterfall attributes the difference to caught value (0.640 against 0.590) and hop holds. At five times the frozen budgets and the six-analyst team bind: every rule catches 0.09 to 0.16 before settlement, losses rise to 0.37 to 0.59, and the probability-space capacity-aware policy fails through its cap again (1{,}405 alerts a day against 900, expiry 0.56). These statements hold at the sampled multipliers only. The multiplier changes how many accepted-MED-like cases occur, not how unreported fraud differs from reported fraud; whether a larger team restores the ordering at five times was not tested; and no level is a claim about actual prevalence or a general domain of applicability.

\subsection{False blocks}
\label{sec:falseblocks}
Under budget threshold and the capacity-aware policy every false block is an analyst decision; only threshold plus block adds score-rule blocks. In the baseline it refuses $404 \pm 26$ legitimate transfers in the 20-day evaluation window, $43 \pm 22$ by the score rule (harm $3{,}200 \pm 1{,}700$ reais) and $361 \pm 20$ by analysts ($28{,}700 \pm 2{,}800$ reais), against $240 \pm 16$ and $302 \pm 16$ analyst blocks for budget threshold and capacity-aware: at the reference scorer the score rule is a small part of the harm. They fall at half capacity (S4: 165 for budget threshold, 127 for threshold plus block, of which 43 by the score rule) because fewer alerts reach a human; counts for every scenario are in the repository. Under drift, over its 70-day window, they are 12.6 and 23.0 a day for budget threshold and threshold plus block, against 12.0 and 20.2 in the baseline, 5 to 14\% higher: the cost of a scorer trained on a warm-up period whose fraud mix has moved.

\section{Limitations}
\label{sec:limitations}

\paragraph{Structural validity.} Receiver-internal features (initial balance, own transfer rate, in-degree) reach single-feature AUC above 0.95 on victim transfers because the dormant activity factor calibrated to zero: simulated mules do little but receive and forward. These features are excluded from the scorer, and the simulator should not be used for receiving-side detection. A development-scale test that recruits mules from legitimate accounts with ordinary activity and balances removes the separability (in-degree AUC 0.99 to 0.86, initial balance 0.97 to 0.68) and leaves the hop-held share almost unchanged ($0.41$ to $0.38$), but raises MED recovery to 0.79, an order of magnitude above the BCB's, because an ordinary balance is what the root block finds. Neither configuration validates mule activity, balances and recovery at once. One reading is that the published decomposition requires mules that are active but near-empty; it is a hypothesis supported by the two points tested, not a demonstrated property. Every policy result in this paper, the hop-held values in particular, is conditional on the current mule and balance model until this is resolved. The journal revision will take, in this order: (1) a factorial screen of mule activity and available balance, with a target-compatibility criterion fixed in advance and every cell reported, followed by the capacity contrast at three, four and six analysts on the compatible configurations; (2) a scenario analysis of analyst service time and review accuracy; (3) single-mechanism perturbations of the frozen 2025 model against the February--April 2026 aggregates, to learn which changes are consistent with the misfit without fitting it; (4) partial adoption of the policy across receiving institutions. Two mechanisms are simplified. Since Resolução BCB~506 (in force from 30~September 2025~\cite{bcb_res506}) a participant that accepts an infraction notification, or marks its own client, must reject every Pix of that user and account, as payer or receiver, except returns (art.~89, \S2), so that a marked mule can neither receive new victims nor forward; the simulator does not model this second interdiction channel, which was in force for the last quarter of the 2025 calibration window and all of 2026, and its effect on the recovery decomposition, on the mule-reuse anchor and on the 40\% marker-hit share below is not characterised; and cautelary blocks are an accounting route drawn with a calibrated probability rather than a lock on the rail, so the forwards of value they cover remain in the rail log, where the policy can score them, and the route overlaps with hop holds; under tracing this lets a layer block double count value a cautelary return already covers, an overstatement of the MED route measured at a quarter of full scale as at most 0.019 of its recoveries (Section~\ref{sec:s8}). The analyst queue's service time and review accuracy are assumed (Section~\ref{sec:queue}), so the capacity results are conditional on them.

\paragraph{Calibration.} The calibrated values are one fitted configuration under the assumed structure; balance, activity, reporting and burning could in other combinations produce the same aggregates, and a local sensitivity of the findings to the calibrated values is left to the journal version. The 2025 recovery rate is overshot by 0.006, inside the 0.02 application tolerance; Section~\ref{sec:regime} gives the scale of a uniform recovery error on loss levels, but the effect of structural calibration error on the policy comparisons is not bounded. February--April 2026 is not reproduced and no parameter is fitted to it; the series shows a change in the composition of accepted cases over the same months, which the fixed fraud population cannot represent and whose cause the aggregates do not identify. The scenario mix, the hub batching rule, the harm cost, the top-up dwell time, chain depth and the production graph depth have no public source and are registered; S8 gives the sensitivity range of the last two.

\paragraph{Deployment.} The policy is applied rail-wide, so hop holds are an effect of universal adoption, not of one PSP's deployment, and partial adoption is not simulated. The Febraban 2026 survey~\cite{febraban2026} and every BCB document in the archive were searched for a figure on fraud value prevented before settlement; none exists, so absolute levels are unanchored and, for the reasons above, optimistic. The scorer is a synthetic one trained on simulated features; the benchmark's result is the ordering along the scorer-quality curve and the range at moderate AUC, not the level at the reference scorer.

\paragraph{Empirical uncertainty.} Fraud is drawn at the accepted-MED incidence, and accepted cases may differ from unreported or rejected ones in size, type, reporting delay and recoverability; S11 changes the rate of the same accepted-MED-like population and does not address that selection. About 40\% of victim transfers reach a key already carrying a fraud marker, with no public anchor; from September 2026 markings are contestable and removable, which may lower real marker prevalence and make it more dynamic than the never-cancelled markers modelled here. Every calibration target is a monthly aggregate; transaction-level validation needs an institutional pilot.

\section{Conclusion and the control problem ahead}
\label{sec:conclusion}

The paper's claims sit at five levels of support, which the reader should keep apart. Calibrated: the frozen model reproduces the 2025 MED recovery level and decomposition within the stated deviations. Observed failure: the February--April 2026 window is not reproduced and was not fitted. Projection: the tracing regime has no published target and depends on the assumed chain depth and graph structure. Conditional benchmark result: within the calibrated mule and forwarding model, the assumed analyst model and rail-wide adoption, the relative performance of the tested rules depends strongly on analyst capacity. Exploratory: the ordering reversal at four analysts, which ten independent seeds did not confirm.

PixSim models Brazil's Pix rail with irreversible settlement, recovery on the MED clock under three regimes, and interdiction through a three-action policy under a finite analyst queue and a regulatory hold window. Four parameters are calibrated to 2025 observables under the 2025 rules and verified on held-out seeds; with the model frozen, full-scale runs reproduce the 2025 recovery rate within 0.006 and its decomposition within 0.02, do not reproduce a 2026 window over which the published series shows a composition change, and project the tracing regime that has no statistic yet. Within the calibrated mule and forwarding model and under universal adoption, post-settlement recovery is strongly constrained by dispersal speed; staffing by arrival profile cuts the fixed rule's alert expiry from 52\% to 2\% at constant hours; halving the analyst team removes the advantage of a fixed threshold-and-block rule over a queue-aware one on twenty paired seeds, on fraud loss and on loss plus harm, while a reversal at two thirds of the team suggested by an exploratory sweep was not confirmed on independent seeds; and the evaluated synthetic scorer at held-out AUC 0.82 reduces lost value by about a quarter relative to no intervention.

The capacity dependence is the reason the simulator was built: the setting is a continuous-time control problem with an irreversible action, a decision deadline, a random recovery clock and analyst capacity as a state, whose objective is expected lost value plus harm, and whose optimal policy is the subject of a companion paper that uses PixSim as its testbed. The rail, clock and queue modules are separated so that another instant rail's recovery rules could replace the MED's, but no such port is demonstrated here.

\section*{Code and data availability}
PixSim is released under the Apache-2.0 licence at \url{https://github.com/BashirZeimarani/pixsim_public}, version 1.0.0, archived at \url{https://doi.org/10.5281/zenodo.22948895}, with the summary CSVs behind every table (\texttt{results/summary/}) and the exact configuration file of every published run; the per-part outputs of the 155 full-scale parts and twelve regime cells are available from the corresponding author on request, and any part can be regenerated from its configuration and seed. Python~3.13 with pinned numpy, pandas, scipy, networkx, scikit-learn, pyarrow and PyYAML; the environment builds from the conda-forge and pip lock files in the repository, the test suite runs under pytest in under a minute, and a reduced-scale end-to-end configuration reproduces the pipeline in minutes on a workstation. \texttt{data/raw} holds every BCB series and document used, with a sources index and retrieval dates; \texttt{docs/assumptions.md} is the parameter register, \texttt{docs/med\_rules.md} the regulatory chronology and regime mapping, \texttt{docs/metrics\_definitions.md} the metric definitions, \texttt{docs/reproducibility.md} the environment checks, and \texttt{CHANGELOG.md} records for every parameter whether it was calibrated or emerged and every correction made. \texttt{BENCHMARK.md} lists configurations, seeds and the scripts that produce every table.

\section*{Use of AI tools}
The simulator, its validation scripts and its test suite were implemented with the assistance of an agentic coding tool (Claude Code, Anthropic) under the corresponding author's direction; Grammarly and Claude were used for grammar checking and editing the text. The model structure, every parameter and its source, the calibration targets, the experimental design, the interpretation of the results and the published text are the authors' own work and responsibility.

\bibliographystyle{unsrtnat}
\bibliography{refs}

\appendix
\section{The rank-space capacity-aware policy}
\label{app:rank}

The rank-space variant alerts the top $k$ share of the scores seen in the current and previous clock hour, with $k = k_0/(1+0.25\,Q/N)$ for queued alerts per analyst $Q/N$ and $k_0$ the operating share at the common queued load; it has no probability cap (A4.13; the slope and the 1024-bin histogram are assumptions). Table~\ref{tab:rank} compares it with the probability-space policy. With a good scorer it is worse on average in every scenario, by $0.012$ $[0.005, 0.019]$ in the baseline and $0.085$ $[0.074, 0.096]$ at half capacity (paired), and on four of five seeds under drift, because it alerts a fixed hourly share rather than a fixed probability and expires more. Its advantage appears only when the scorer saturates: it holds through $\sigma=12$ where the probability-space policy has collapsed (Table~\ref{tab:s7}). For a benchmark of fixed rules the probability-space policy is the reference; a deployed system would need the cap removed by construction.

\begin{table}[htbp]
\centering\small
\caption{Loss share of the rank-space capacity-aware policy against the probability-space one, mean~$\pm$~sd, with the paired difference (rank minus probability) and its 95\% percentile bootstrap interval (10{,}000 resamples of paired seeds); $^{\dagger}$for the five-seed scenarios, the range of the per-seed differences (positive on four of five seeds in S2, on all five in S5).}
\label{tab:rank}
\begin{tabular}{lcccc}
\toprule
Scenario & $n$ & Probability-space & Rank-space & Paired difference \\
\midrule
S1 baseline & 20 & \pmx{0.221}{0.014} & \pmx{0.233}{0.017} & $+0.012$ $[+0.005, +0.019]$ \\
S2 drift & 5 & \pmx{0.195}{0.007} & \pmx{0.211}{0.011} & $+0.016$, $-0.000$ to $+0.025^{\dagger}$ \\
S4 half capacity & 20 & \pmx{0.314}{0.017} & \pmx{0.399}{0.019} & $+0.085$ $[+0.074, +0.096]$ \\
S5 flat staffing & 5 & \pmx{0.267}{0.027} & \pmx{0.319}{0.034} & $+0.052$, $+0.002$ to $+0.083^{\dagger}$ \\
S6 no DICT & 10 & \pmx{0.315}{0.021} & \pmx{0.331}{0.021} & $+0.016$ $[+0.003, +0.029]$ \\
S9 MED~2.0 no tracing & 10 & \pmx{0.204}{0.013} & \pmx{0.218}{0.015} & $+0.014$ $[+0.009, +0.019]$ \\
\bottomrule
\end{tabular}
\end{table}

\section{Runs behind the tables}
\label{app:runs}

\begin{table}[htbp]
\centering\scriptsize
\caption{Full-scale runs behind every table: configuration, its hash, seeds and completed parts (all complete). Every part runs the five reference policies unless stated. A part shared by several tables is counted once: 155 distinct parts, plus twelve MED-clock validation cells without interdiction.}
\label{tab:runs}
\setlength{\tabcolsep}{4pt}
\begin{tabular}{>{\raggedright\arraybackslash}p{2.9cm}>{\raggedright\arraybackslash}p{5.0cm}p{1.6cm}p{3.0cm}r}
\toprule
Tables & Scenario (configuration) & Hash & Seeds & Parts \\
\midrule
\ref{tab:sixconfig}, \ref{tab:paired}, \ref{tab:waterfall}, \ref{tab:rank}; S7, S11, S10 reference & S1 baseline (\texttt{fullscale}) & e4a21c75 & 20260910--20260929 & 20 \\
\ref{tab:sixconfig}, \ref{tab:paired}, \ref{tab:rank} & S2 drift (\texttt{fullscale\_drift}) & 8c80a08d & 20260910--20260914 & 5 \\
\ref{tab:sixconfig}, \ref{tab:paired}, \ref{tab:capsweep}, \ref{tab:waterfall}, \ref{tab:rank} & S4 half capacity (\texttt{fullscale\_halfcap}) & cb8130c9 & 20260910--20260929 & 20 \\
\ref{tab:sixconfig}, \ref{tab:paired}, \ref{tab:waterfall}, \ref{tab:rank} & S5 flat staffing (\texttt{fullscale\_flat}) & 8533dca9 & 20260910--20260914 & 5 \\
\ref{tab:sixconfig}, \ref{tab:paired}, \ref{tab:waterfall}, \ref{tab:s7}, \ref{tab:rank} & S6 no DICT (\texttt{fullscale\_nodict}) & b487ff4a & 20260910--20260919 & 10 \\
\ref{tab:sixconfig}, \ref{tab:paired}, \ref{tab:rank} & S9 MED~2.0 no tracing (\texttt{fullscale\_s9\_med2\_notrace}) & 31e67de2 & 20260910--20260919 & 10 \\
Section~\ref{sec:s8} & S8 tracing, 6 configurations (MaxHops $\{1,2\}$ $\times$ depth $\{1.5,2,3\}$; 2 policies) & 6 hashes & 20260910--20260912 & 18 \\
\ref{tab:s7} & S7 scorer quality, 6 configurations (noise $\sigma \in \{3,6,8,12,24\}$, no amounts) & 6 hashes & 20260910--20260912 & 18 \\
\ref{tab:capsweep} & S10 four analysts (\texttt{fullscale\_cap4}) & f29a8ae4 & 20260910--20260919 & 10 \\
\ref{tab:capsweep} & S10 five analysts (\texttt{fullscale\_cap5}) & 99415b5c & 20260910--20260919 & 10 \\
\ref{tab:capsweep} & S10c confirmation, four and five analysts & f29a8ae4, 99415b5c & 20260920--20260929 & 20 \\
Section~\ref{sec:s11} & S11 prevalence, 3 configurations ($0.5\times$, $2\times$, $5\times$) & 3 hashes & 20260910--20260912 & 9 \\
\ref{tab:regimes} & MED-clock validation cells, three regimes, separate and common random numbers & 2b4ca787 & 20260910--20260912 & 9 + 3 \\
\bottomrule
\end{tabular}
\end{table}

The per-configuration hashes of S7, S8 and S11 are listed in the repository's benchmark file.

\end{document}